\documentclass[11pt]{article}

\usepackage{microtype} 
\usepackage{booktabs}  
\usepackage{url}  

\usepackage{amsmath}
\usepackage{amsthm}
\usepackage[utf8]{inputenc} 
\usepackage[T1]{fontenc}    
\usepackage{amsfonts}       
\usepackage{nicefrac}       
\usepackage{lipsum}
\usepackage{fancyhdr}       
\usepackage{graphicx}       
\usepackage{amsmath,amssymb} 
\usepackage{caption}
\usepackage{subcaption}
\usepackage{xcolor}         
\usepackage{color}
\usepackage{multirow}
\usepackage[final]{hyperref} 

\newcommand{\features}[3]{\langle #1, #2, #3 \rangle}

\hypersetup{
    colorlinks=true,
    linkcolor=red,
    citecolor=blue,
    filecolor=magenta,
    urlcolor=cyan,
}

\usepackage[finalworkshop]{automl}

\usepackage{natbib}

\title{How to Learn and Represent Abstractions: \\ An Investigation using Symbolic Alchemy}

\author[1]{\nameemail{Badr AlKhamissi}{badr@khamissi.com}}
\author[1]{Akshay Srinivasan}
\author[2]{Zeb-Kurth Nelson}
\author[2]{Sam Ritter}

\affil[1]{Sony AI}
\affil[2]{DeepMind}

\begin{document}

\maketitle

\begin{abstract}

    Alchemy is a new meta-learning environment rich enough to contain interesting abstractions, yet simple enough to make fine-grained analysis tractable. Further, Alchemy provides an optional symbolic interface that enables meta-RL research without a large compute budget. In this work, we take the first steps toward using \textit{Symbolic Alchemy} to identify design choices that enable deep-RL agents to learn various types of abstraction. Then, using a variety of behavioral and introspective analyses we investigate how our trained agents use and represent abstract task variables, and find intriguing connections to the neuroscience of abstraction. We conclude by discussing the next steps for using meta-RL and Alchemy to better understand the representation of abstract variables in the brain.

\end{abstract}

\section{Introduction}

Humans display the remarkable ability to use abstractions to guide their behavior. They are able to abstract over the sensorimotor details of a situation to derive the general principles involved, and to use those principles to behave effectively. A classic example is the use of abstract knowledge to guide behavior in a restaurant; for example, people know that they must pay at the end, even though they have never seen that exact restaurant or paid for that specific meal \citep{wallis_single_neurons_2001}. Neuroscientific studies have made progress in understanding how the brain might represent abstract rules \citep{mansouri_review_2020}, establishing that prefrontal cortex plays a significant role \citep{milner1963effects}, with more recent work implicating the hippocampus as well \citep{samborska_complementary_2021}. One limitation of these studies is that they use extremely simple tasks, where optimal behavior can be derived as a nearly trivial function of the inputs. As a result, there is a large gap between the abstractions studied in the experiments and the richness at play in human life.

A recent development presents an opportunity to begin to bridge this gap. The Alchemy benchmark was proposed by \cite{wang_alchemy_2021} to make fine-grained analysis and interpretation of meta-RL agents possible while maintaining the complexity needed for more interesting conceptual abstractions to be learnt. In the present work, we use a biologically inspired model on the symbolic version of the Alchemy benchmark \citep{wang_alchemy_2021} to investigate the learning and representation of more complex abstractions than those studied previously in neuroscience. Our model builds on the meta-RL framework of \citet{wang_pfc_2018} who showed that by considering the prefrontal cortex (PFC) as its own meta-RL system, that is driven by dopamine-based synaptic learning, one can account for a wide range of behavioral and neurophysiological findings. Our model builds on that by extending their core recurrent network with an episodic memory via a modified transformer block to represent the hippocampus, similarly to \citep{ritter18episodic}.

The main contributions of the paper can be summarized as follows: (1) we show a way in which researchers can achieve high performance in Symbolic Alchemy -- albeit with some tricks -- without having access to vast computational resources as is usually required for deep-RL research. To be specific, all experiments done for this paper were run on a single-GPU (Tesla T4) machine. (2) We propose a hypothesis based on empirical results into why previous agents failed to solve Symbolic Alchemy despite being much more powerful than the one used in this work. (3) We release a tool for visualizing the chemistry and latent space of any given episode in Symbolic Alchemy to better help researchers debug the behavior of their agents. (4) We present a new kind of behavioral analysis that can be done on Alchemy to test whether the agent succeeded or failed in acquiring specific pieces of abstract knowledge. (5) Finally, we demonstrate that, just as in animals \citep{wallis_single_neurons_2001, wallis_2003, rahmat_abstract_rules_2006}, single-units of the LSTM and transformer encode abstract task variables. Moreover, single-unit analyses revealed evidence for distinct functional roles for LSTM and transformers units. We draw a connection between this observation and the differential roles of PFC and hippocampus observed in recent neuroscience experiments \citep{samborska_complementary_2021}.

\section{Methods}

    \subsection{Symbolic Alchemy}
    \label{sec:task}
    
    
    The Alchemy benchmark was proposed by \cite{wang_alchemy_2021} to make fine-grained analysis and interpretation of meta-RL agents possible while maintaining the complexity needed for interesting conceptual abstractions to be learnt. Unlike other meta-RL task distributions, Alchemy's accessibility allows us to compare our model's performance against a Bayesian learner referred to as an `Ideal Observer'. \cite{wang_alchemy_2021} also develop a `Random Heuristic' that we use as a reference for evaluating our agent's understanding of specific abstract principles. The task itself is divided into episodes, each of which consists of $10$ trials. In our experiments, the number of timesteps per trial was set to $15$ to speedup training.
    
    \paragraph{Stones and Potions} The goal of the agent within each trial is to transform three stones into a more valuable form -- the value of which is tied to the stone's perceptual features -- by applying a sequence of different potions on each of them. The agent can then collect the reward associated with a stone by dropping it into a central cauldron. The stone's appearance can change along only one of three dimensions at a time: size, color and shape. Each potion on the other hand is characterized by one of $6$ hues that dictates the transformative effect it has on a specific stone according to some `chemistry' that is sampled from a structured generative process at the beginning of each episode. There are $12$ potions in each trial, each of which is consumed (i.e. can not be re-used) once applied on a stone.
    
\begin{figure}[h]
        \centering
        \begin{subfigure}{0.25\textwidth}
            \includegraphics[width=1\linewidth]{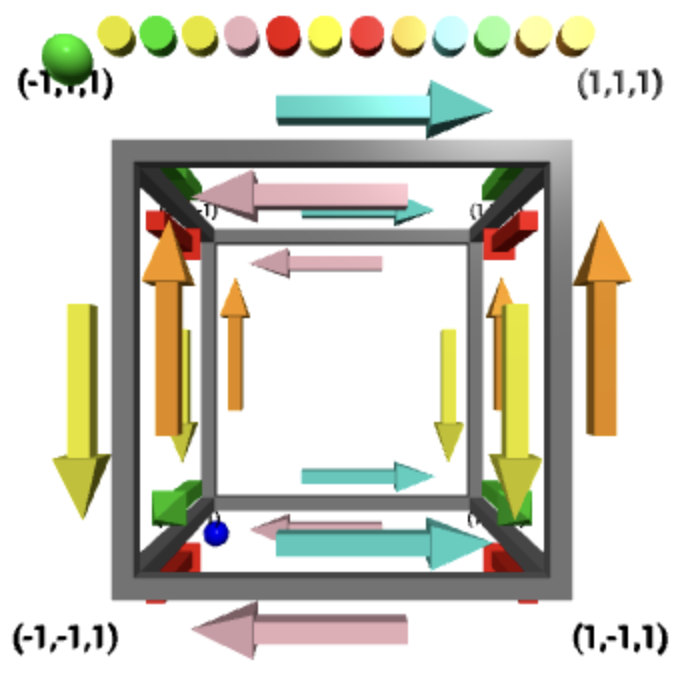}
        \end{subfigure}%
        \begin{subfigure}{0.25\textwidth}
            \includegraphics[width=1\linewidth]{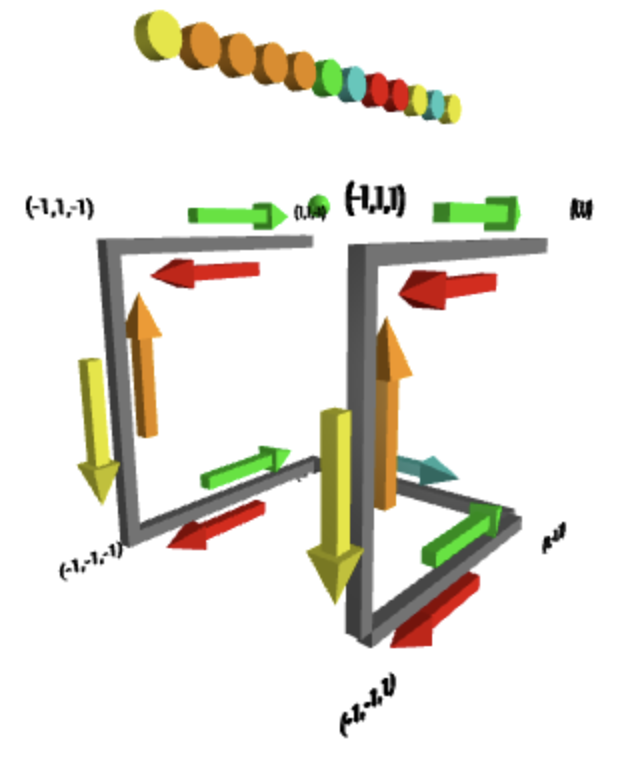}
        \end{subfigure}%
        \begin{subfigure}{0.25\textwidth}
            \includegraphics[width=1\linewidth]{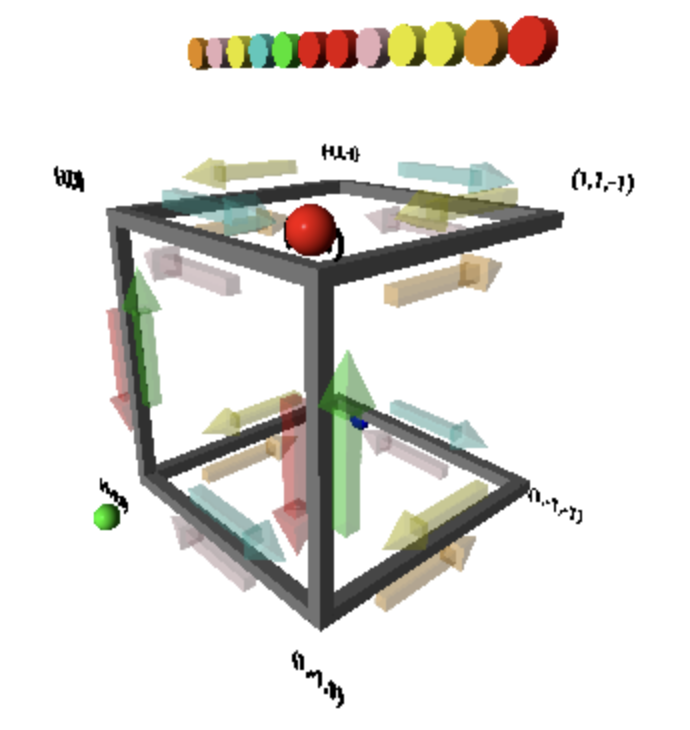}
        \end{subfigure}%
        \begin{subfigure}{0.25\textwidth}
            \includegraphics[width=1\linewidth]{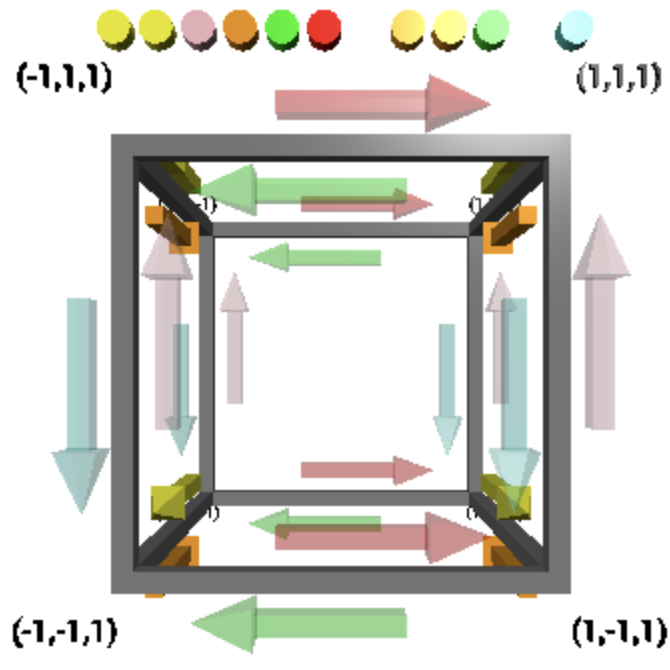}
        \end{subfigure}
        \caption{Visualizations of chemistries created using the Symbolic Alchemy Visualization Tool, which we are releasing for public use. The coordinates on the cube's vertices indicate the latent state and thus reward of the stone in that position. Translucent arrows indicate that the agent has not yet discovered the effect of that color. The stones are represented by a red, blue and green spheres. The available potions are shown floating above the cube with their corresponding color, and when a potion is consumed it will no longer be visible. In some chemistries, edges of the graph may be missing as can be seen in the 2nd and 3rd snapshots.}
        \label{fig:sa_viz}
    \end{figure}

    \paragraph{Chemistry} The chemistry dictates the causal structure that governs each episode as well as the possible set of stone perceptual features that can occur. It can be visualized as a cube, where each vertex correspond to a specific stone value in the latent space, which can be one of (-3, -1, +1, +15), or a specific appearance in the perceptual space. The potion effects run along the edges of the cube as shown in Figure \ref{fig:sa_viz}. Edges can be missing, creating a bottleneck that the agent must pass through to reach a high rewarding state. This may require passing through intermediate lower value states first.
    
    \paragraph{Abstract Principles} There are certain rules and constraints that span across episodes which makes Alchemy a meta-learning benchmark. The agent is expected to learn to identify and exploit those regularities in order to achieve high performance in the task. This includes: 
    
    \begin{itemize}
        \item \textbf{Consistency}: Potions of the same color will always have the same effect on stones with the same visual features within an episode. 
        \item \textbf{Parallelism}: Each potion color has the same direction of effect regardless of the other features (e.g. a red potion will always change the color of a stone from blue to purple irrespective of the size and shape of that stone).
        \item \textbf{Missing Edges}: Overlaid on that parallelism, some edges can be disabled for an episode. Therefore once discovered the agent shouldn't attempt to traverse that missing edge.
        \item \textbf{Potion Pairs}: Potions come in pairs with opposite effects (red/green, yellow/orange, pink/turquoise). The agent should know those pairs since they are consistent across all episodes. For example, if the effect of the orange potion is to increase the size of a stone, then the effect of the yellow potion will be to decrease the size of the stone.
    \end{itemize}

    \paragraph{State and Action Spaces} In the input representation, each stone is represented by its three perceptual features, its reward in the current latent state and whether it has been deposited into the cauldron or not. For the potions, we use a modified representation from the one proposed by \cite{wang_alchemy_2021}; instead of representing the state of each of the $12$ potions as elements of a single vector, we represent the remaining number of potions per hue. Therefore the state space is comprised of a $21$-dimensional vector. The action space is also modified in a similar manner, where the agent chooses which color to apply to which stone, or whether to deposit a specific stone to the cauldron. When a specific stone-color combination is chosen, a wrapper then randomly selects one of the available potions with that color and apply it to the selected stone. Alternatively a \texttt{no-op} action can be chosen, which makes the number of possible actions $22$. 
    
    \paragraph{Rewards and Penalties} In addition to the rewards the agent receives whenever depositing a stone into the cauldron, we found that penalizing the agent in three specific scenarios sped up the rate of convergence considerably. Specifically, we gave a reward of $-0.2$ if the action taken results in a null transition (i.e. does not have any effect on the stone) but given that it is not a \texttt{no-op} action. In the second case, the agent was penalized with a reward of $-1$ whenever it chose to use a potion hue that is not available (i.e. it should learn to never use a hue when its corresponding entry is zero in the input representation) or a stone that has already been cached in the cauldron. Finally, an additional penalty of $-1$ was given when the agent chose the same potion color consecutively on the same stone.

    \subsection{Visualizing the Latent Space}
    \label{sec:latent_space_viz}
    In order to simplify the process of qualitatively debugging the agent's behavior, we developed and are releasing\footnote{\url{https://github.com/bkhmsi/symbolic\_alchemy\_visualizer}} a tool for Symbolic Alchemy that visualizes the topology of the underlying causal graph for a given episode. Overlaid on it are the positions of the stones in the latent space along with arrows indicating the direction of effect of the available potions. The Ideal Observer's belief about a particular potion color or edge is also indicated using the opacity of the corresponding object. That way we can evaluate the agent's actions with respect to the belief state of an agent that has perfect understanding of the structure of the task. 
    
    \subsection{Agent Architecture}
    \label{sec:agent_arch}
    
    \begin{figure}
        \centering
        \includegraphics[width=1\linewidth]{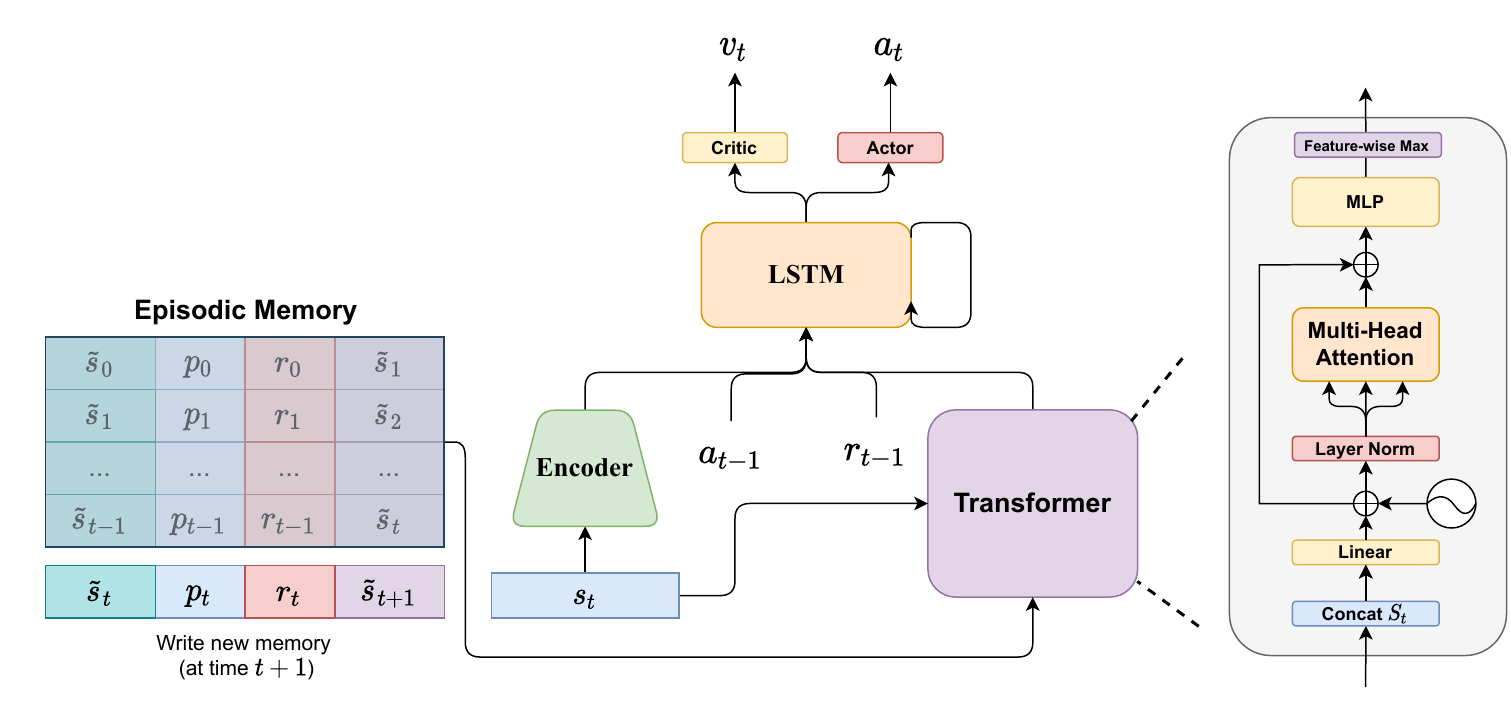}
        \caption{In the center of the architecture is an LSTM that takes as input the action and reward of the previous timestep, an encoded version of the current state, and a distilled representation of the relevant memory entries. The encoder is just a two-layered MLP. The episodic memory stores the features of the stone being transformed $\widetilde{s}_t$, a one-hot encoding of the applied potion color $p_t$, the reward $r_t$, and the resultant features of the stone after transformation $\widetilde{s}_{t+1}$. The transformer block architecture is fully described in Section \ref{sec:agent_arch}. The output of the LSTM is finally given to the policy and value networks, each of which is a linear layer, to generate an action and the value estimate respectively. } 
        \label{fig:arch}
    \end{figure}
    
    We use a biologically inspired architecture that maps to functions and neural structures in the brain. Specifically, we build on the work of \cite{wang_pfc_2018} in which the prefrontal cortex (PFC) is conceptualized as forming a gated recurrent neural network (characterized as an LSTM \citep{lstm}) and augment it with an episodic memory which is connected to the LSTM via a single modified transformer block from \cite{ritter_rapid_2020} (see Figure \ref{fig:arch}).  
    
    The LSTM, which has been shown to have analogies with prefrontal gating theories \citep{chatham15gates}, takes as input on each time-step an encoded version of the current state, the reward and action it had taken at the preceding time-step and a compressed representation of the previous memories experienced in the current episode. Since the LSTM learns to distill that information over the course of an episode in its hidden state, it can be considered as a form of working-memory \citep{antonio_pfc_wm_2015}.
    
    On the other hand, the transformer architecture takes as input the current state $s_t$ and the entries of the episodic memory $\{m_i\}_{i=0}^t$ on each time-step. The state $s_t$ is then concatenated to each $m_i$. This matrix is then transformed feature-wise by a shared linear layer, before being passed to the planner module of \cite{ritter_rapid_2020} described using the following equations: $s^*_t = \operatorname{ReLU}(x + \phi(x))$ where $x = \{[m_i, s_t]\}_{i=0}^t$ and $\phi(x) = \operatorname{MHA}(\operatorname{LayerNorm}(x))$ where $\operatorname{MHA}$ is the multi-head dot-product self-attention mechanism described by \cite{vaswani_17} with layer-normalization \citep{Ba2016LayerN} applied to the input. The output is an attended view of the agent's past relevant experience in the episode given the current state. This is then passed to a two-layer shared MLP, the output of which is pooled using a feature-wise max operation. Since the transformer we use is referred to as an Episodic Planner Network (\textbf{EPN}) in \citep{ritter_rapid_2020} we refer to our agent as \textbf{A2C EPN}.
    
    \subsection{Experimental Setup}
    \label{sec:exp_setup}
    
    The agent is trained using the synchronous version of the Advantage Actor-Critic (A2C) RL algorithm \citep{mnih16a3c} with a batch-size of $8$. The value coefficient was set to $\beta_V = 0.5$ and the initial entropy coefficient to $\beta_E=0.1$ which was decayed in a linear fashion throughout training. The encoder is a stack of $2$ affine layers with $32$ units each, and an $\operatorname{ELU}$ non-linearity in-between \citep{elu}. The LSTM has $256$ hidden units, while the transformer block contains $4$ attention heads with a dimensionality of $64$. The MLP is similarly a stack of two affine layers with $64$ units each, with an $\operatorname{ELU}$ non-linearity in-between. The episodic memory operated with a maximum size of $150$ entries, and was reset after each episode. We used the Adam optimizer \citep{adam} with an initial learning rate of $7.5e-4$. The learning rate was decayed linearly from the start of training. The gradient was clipped at a maximum norm of $100$. We found that starting with a small discount factor of $\gamma=0.7$ worked best. The resultant model was then finetuned using a higher value ($\gamma=0.95$) after convergence. The number of unroll steps is $20$. The code is made open-source\footnote{\url{https://github.com/bkhmsi/alchemist}}.

\section{Results}

    The results and analyses presented in this section have all been done on the $1000$ held-out evaluation episodes provided by the benchmark using the agent described in Section \ref{sec:agent_arch}.  
    
    \subsection{Performance}
    

    The A2C EPN agent achieves an average episode score of $272.85$ with the modified representations, which is significantly higher than the much more powerful VMPO agent with a gated transformer XL network \citep{Song2020VMPO, parisotto_19} used by \cite{wang_alchemy_2021}. In an attempt to identify which aspects contributed to this leap in performance, we re-trained our agent but using the canonical representation in each of the input, output and memory separately as shown in Table \ref{tab:perf_comparison} (more details about each in the Appendix \ref{sec:io_explanation}). The results show a significant drop to almost the same performance reported by \cite{wang_alchemy_2021} when using the canonical action space. Interestingly, evaluating the same agent but without committing anything into the episodic memory led to the same performance (see Score w/o Memory column in Table \ref{tab:perf_comparison}), indicating that the model was not making use of its long-term history to appraise the value of future actions. 
    
    \begin{figure}
        \centering
        \begin{subfigure}{1.0\textwidth}
            \includegraphics[width=1\linewidth]{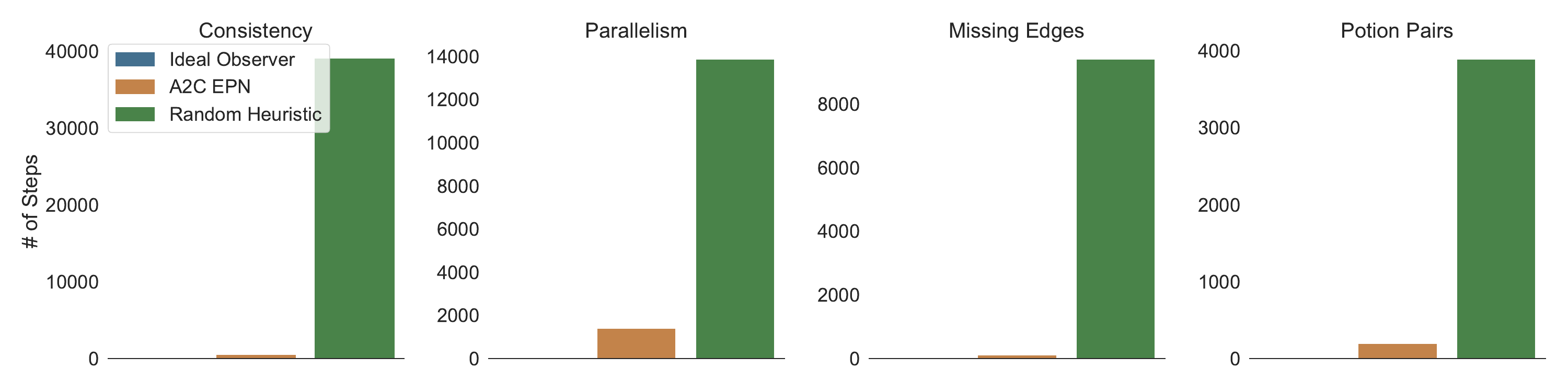}
        \end{subfigure}
        \begin{subfigure}{0.3\textwidth}
            \includegraphics[width=1\linewidth]{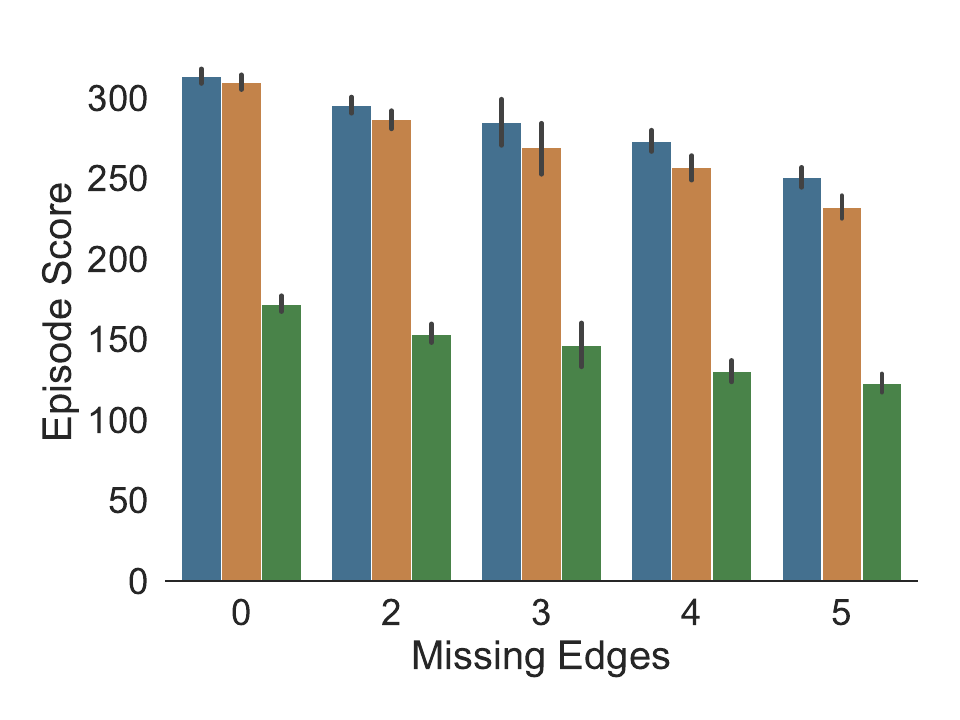}
        \end{subfigure}%
        \begin{subfigure}{0.7\textwidth}
            \includegraphics[width=1\linewidth]{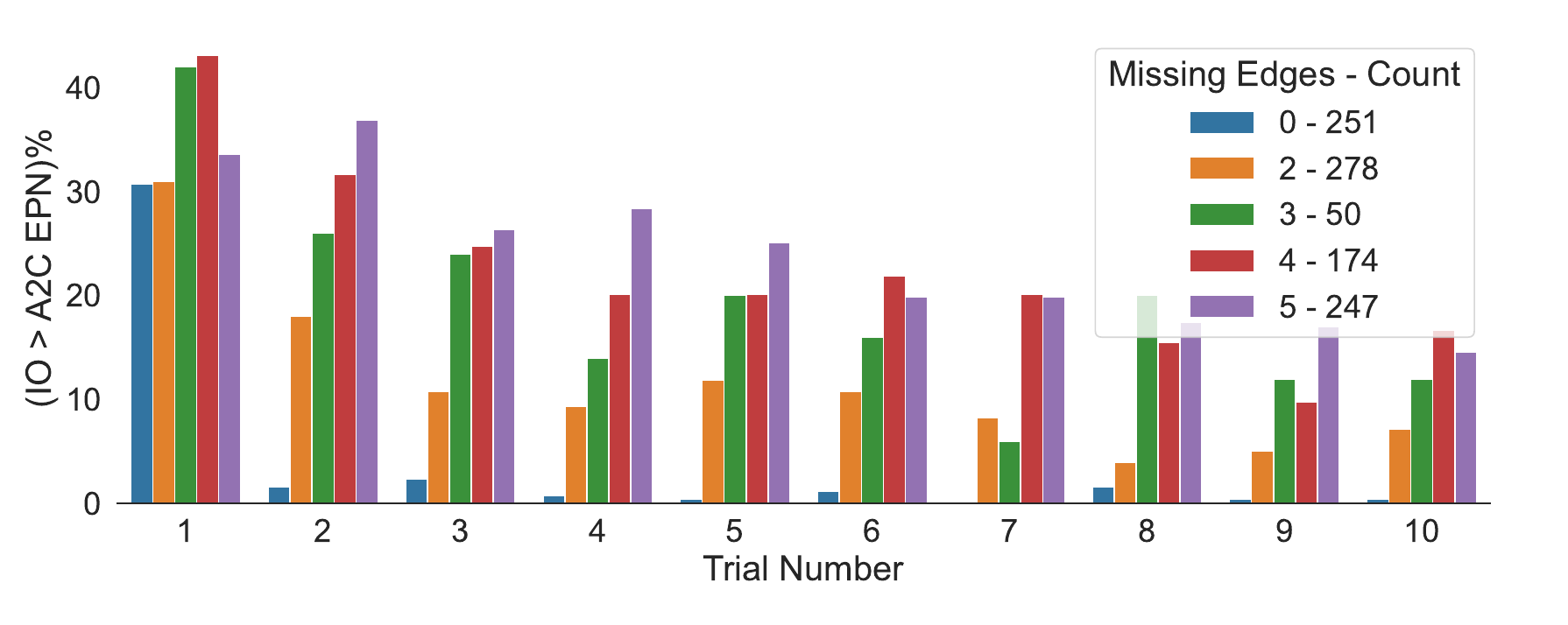}
        \end{subfigure}
        \caption{\textbf{Top}: Behavioral results showing the agent's lack of understanding of several abstract principles that span across episodes. In each plot, we report the number of times the agent took a null transition that it should have avoided if it understood the corresponding concept. The Ideal Observer is zero in all cases demonstrating perfect understanding of each principle. Our agent performs significantly better than the Random Heuristic but still takes a lot of actions where it could know better in the Parallelism and Potion Pairs analyses. \textbf{Bottom-left}: The average episode score as a function of the number of missing edges for each agent. \textbf{Bottom-right}: The percentage of episodes where the Idea Observer scores more than the A2C EPN agent at a given trial.}
        \label{fig:results}
    \end{figure}

    \begin{table}[h]
    \centering 
    \caption{Evaluation episode scores comparing the effect of the canonical representation (indicated by an `o') proposed by the Alchemy benchmark to the modified one (indicated by an `x') described in Section \ref{sec:task}. The VMPO result taken from \citep{wang_alchemy_2021}. The no bottleneck column indicate the average score of the evaluation episodes with no missing edges.}
    \begin{tabular}{lcccccc}
    \hline
    \textbf{Agent}                    & \textbf{Input} & \textbf{Output} & \textbf{Mem} & \textbf{Score $\pm$ SEM} & \textbf{\begin{tabular}[c]{@{}c@{}}Score \\ (w/o Mem)\end{tabular}} & \textbf{\begin{tabular}[c]{@{}c@{}}No \\ Bottleneck\end{tabular}} \\ \hline
    \multirow{4}{*}{\textbf{A2C EPN}} & x              & x               & x               & 272.85 $\pm$ 1.78        & 168.88                                                               & 309.90                                                            \\
                                      & o              & x               & x               & 243.83 $\pm$ 2.21        & 171.59                                                                 & 300.79                                                            \\
                                      & o              & o               & x               & 156.34 $\pm$ 1.57        & 156.70                                                              & 181.41                                                            \\
                                      & o              & o               & o               & 158.91 $\pm$ 1.60         & 153.40                                                              & 182.10                                                            \\ \hline
                                      
    \textbf{VMPO}                     & o              & o               & -               & 155.40 $\pm$ 1.60         & -                                                                   & -                                                            \\
        
    \textbf{Random Heuristic}         & -              & -               & -               & 146.07 $\pm$ 1.55         & -                                                                   & 172.11                                                            \\
    \textbf{Ideal Observer}           & -              & -               & -               & 284.42 $\pm$ 1.59        & -                                                                   & 313.31                                                            \\ \hline
    \end{tabular}
    \label{tab:perf_comparison}
    \end{table}
    
    This inability to exploit the memory can be largely attributed to the notion of positional output, where the agent is required to choose an instance of the potion and not the abstract color that causes the transformative effect. Since each potion slot can have more than one color across different trials, the same action will have different effects within the same episode. For instance, action $a_2$ in trial $k$ can be a pink potion while in another trail can correspond to an orange potion. In order to have an appropriate mapping between each instance and its abstract color one will need a more suitable architecture that would be able to recapitulate the information we gave to the system via the custom encoding. The series of experiments we presented show where such an architecture search could begin. Specifically, an architecture design that gets the right information into the memory and provides sufficiently flexible neural networks to process the memories in order to reproduce the information contained in our modified representation. In this work we used the custom encoding since it enabled us to pinpoint where the architectural problem lies and perform interesting analyses that we present in the following sections with a small compute budget.  
    
    To measure the effect of reward shaping on the final performance, we evaluated the A2C EPN agent without giving it any additional rewards or penalties. This achieved a score of $236.06 \pm 2.18$, which suggests that exploration is a bottleneck. In other words, the problem does not really have to do much with learning from examples to represent abstract variables, but instead is just about having a data distribution that's sufficiently broad.
    
    We found it useful as well to compare the results with respect to the number of missing edges in the graph (see Figure \ref{fig:results}), since the A2C EPN agent's performance approached that of the Ideal Observer in the case where there are no bottlenecks. In an effort to bridge this gap, we found that this difference is more pronounced in earlier trials. It is especially visible in the first trial in the case where there are no missing edges as shown in Figure \ref{fig:results}, where the A2C EPN agent performs similar to or better than the Ideal Observer in far more number of episodes after the first trial. This can be mitigated in future work by incorporating more advanced exploration methods.
    
\begin{figure}[h]
        \centering
        \begin{subfigure}{0.5\textwidth}
             \includegraphics[width=1\linewidth]{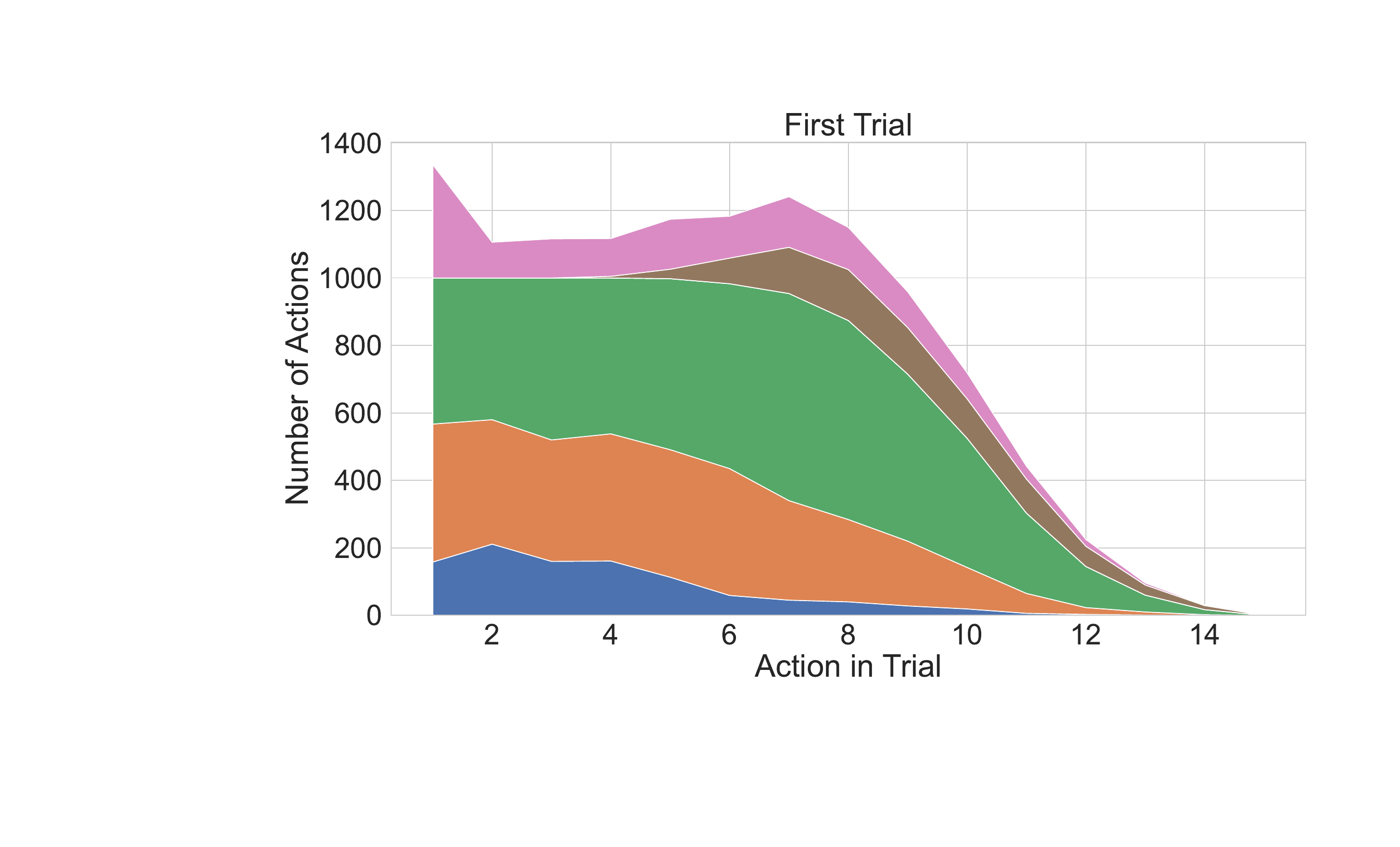}
        \end{subfigure}%
        \begin{subfigure}{0.5\textwidth}
             \includegraphics[width=1\linewidth]{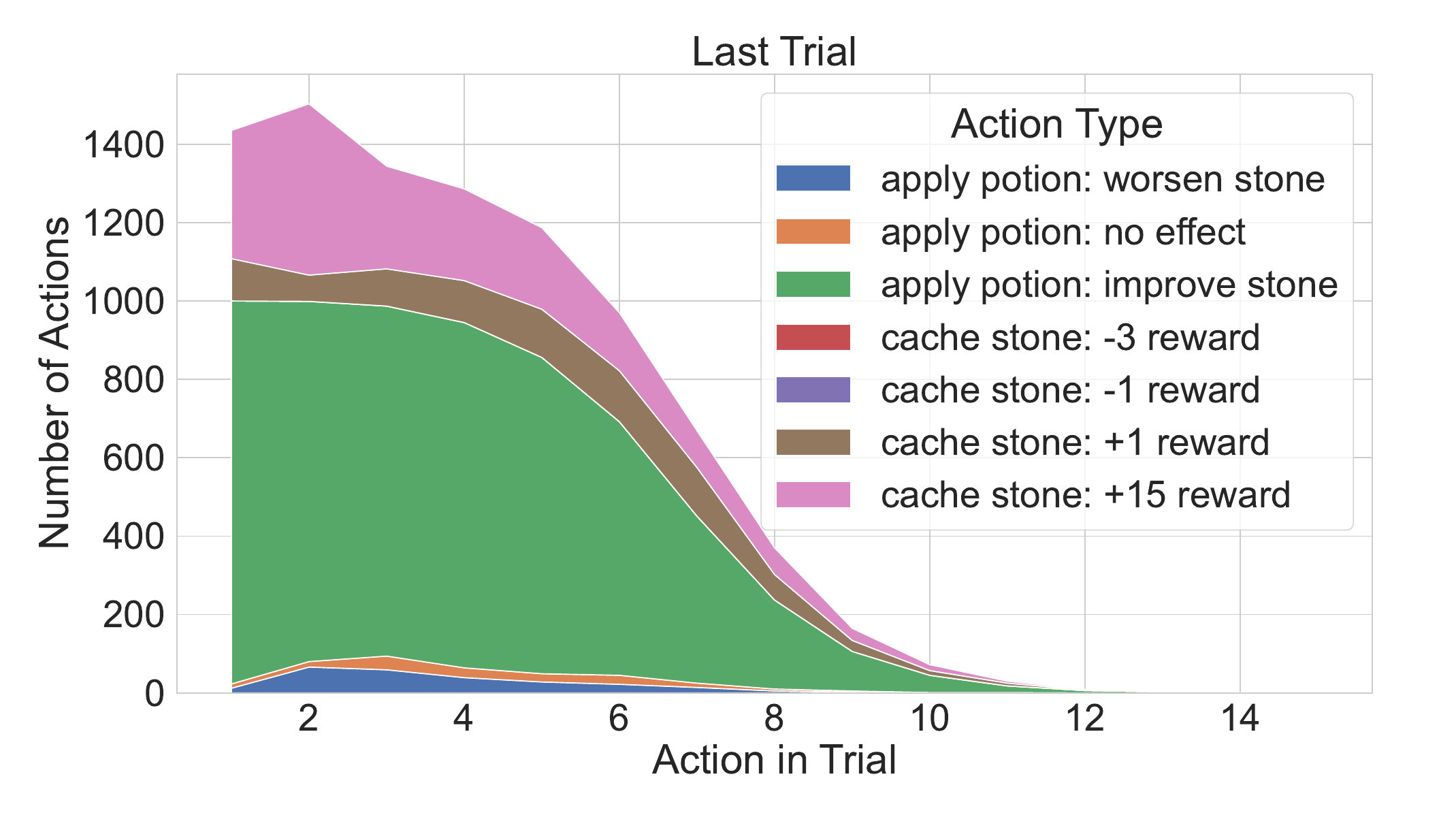}
        \end{subfigure}
        \caption{Comparing different action types throughout the first and last trial in a similar manner to \cite{wang_alchemy_2021}. The A2C EPN agent shows analogous behavior to that of the Ideal Observer (see Figure \ref{fig:io_action_type} in Appendix), indicating an ability to adapt strategies (exploration vs exploitation) throughout the course of the episode.}
        \label{fig:action_type}
    \end{figure}

    \paragraph{Action Types} Similar to the analysis done in \cite{wang_alchemy_2021}, Figure \ref{fig:action_type} shows the number of times, throughout the first and last trial, the agent applied a potion that worsened, improved or had no effect on the value of the stone and the number of times the agent deposited a stone into the central cauldron as a function of its reward. It can be seen that the A2C EPN agent almost never deposits a negative reward stone and more importantly, it adapts its strategy throughout the course of the episode. Concretely, in the first trial it performs a lot of exploratory actions by trying out potions that do not have an effect on the stone (the orange area) while in the last trial it shifts towards a more exploitative strategy by performing more actions that improved the value of the stone as it acquired knowledge about the episode's chemistry. This ability to adapt is indicative of good meta-learning performance and is similar in behavior to what we see from the Ideal Observer as shown in the Appendix.

    \subsection{Behavioral Tests}
    \label{sec:behavioral_tests}
    
    In the following behavioral tests we report the number of times the agent took a null transition (i.e. applying a potion that has no effect on a stone) that wasn't the result of a missing edge or choosing a potion color or stone that was not available, but due to a lack of understanding of a specific abstract principle. We compare our agent's behavior to that of the Ideal Observer and the Random Heuristic. 
    
    To put it formally, let $t: \features{x}{y}{z} \xrightarrow{p} \features{w}{v}{u}$ be a transition that the agent has observed earlier in the episode, where $x$, $y$ and $z$ are the color, size and shape of the stone in question respectively, $p$ is the color of the potion being applied to the stone, and $w$, $v$ and $u$ are the stone perceptual features after applying the potion on it. Note that $\features{x}{y}{z}$ can be equal to $\features{w}{v}{u}$, and we call that a `null transition'. Since there are no rotated graphs in the evaluation episodes, each feature dimension can only take one of two values. This implies that each stone can only be in one of $2^3 = 8$ possible perceptual states. Figure \ref{fig:results} shows the results of each tested abstract principle.
  
    \paragraph{Consistency} Here we count the number of times the agent applied a potion with color $p$ on a stone with features $\features{x}{y}{z}$ after observing that $t$ is a null transition. In other words, we have evidence that the agent does not understand consistency if it applied a potion on a stone after observing that this particular potion color has no effect on a stone with the same visual features earlier in the episode.
    
    \paragraph{Parallelism} In this test, we calculate the frequency in which the agent applied a potion with color $p$ on a stone with features $\features{a}{b}{\bar{z}}$ for the first time after it observed that $t$ is not a null transition, where $\bar{z}$ is the other possible shape, assuming that $p$ is responsible for transforming the shape of the stone. For example, suppose the agent observed a red potion transforming a large blue round stone to pointy. Subsequently in the episode, it should never apply a red potion to any pointy stone, including small or purple stones.
    
    \paragraph{Missing Edges} The agent will show a lack of understanding of missing edges if it re-applied potion $p$ on a stone with the same latent state as the stone with $\features{x}{y}{z}$ after observing that $t$ is a null transition as a result of a missing edge. In other words, we compute the frequency in which the agent attempted to transform a stone after `discovering' that this color has no effect on the current latent state due to a missing edge in the graph. Note that the agent does not have access to this information, but must identify it by experimentation. 
    
    
    \paragraph{Potion Pairs} To test whether the agent demonstrate an understanding that potions come in pairs with opposite effects, we count the number of steps in which the agent made a null transition as a result of applying a potion color $\bar{p}$ that it has not seen the effect of before, but has observed the effect of its opposite color previously in the episode (i.e. $t$ was not a null transition). For example, an agent should know that yellow potions decreases the size of stones after only observing that an orange potion transformed a small stone to a large one (without having to see the effect of a yellow potion before). Therefore, it should never apply a yellow potion to a small stone as to avoid a null transition.  
    
    
    
    \subsection{Single-Unit Activations}
    
    \begin{figure}
        \centering
        \begin{subfigure}{0.25\textwidth}
        \caption{}
            \includegraphics[width=1\linewidth]{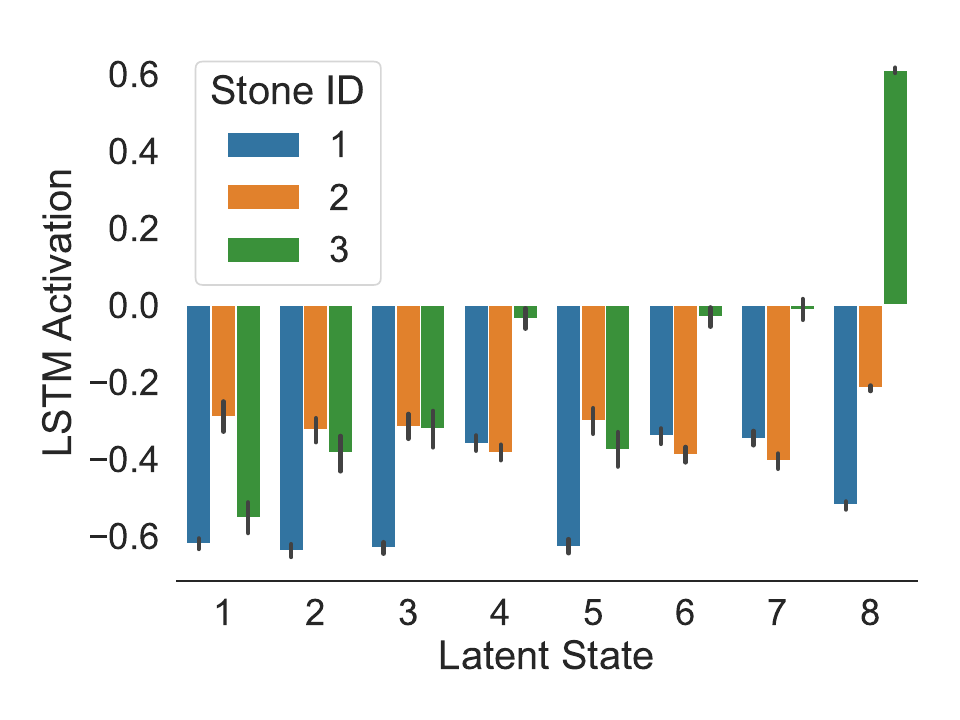}
        \end{subfigure}%
        \begin{subfigure}{0.25\textwidth}
        \caption{}
            \includegraphics[width=1\linewidth]{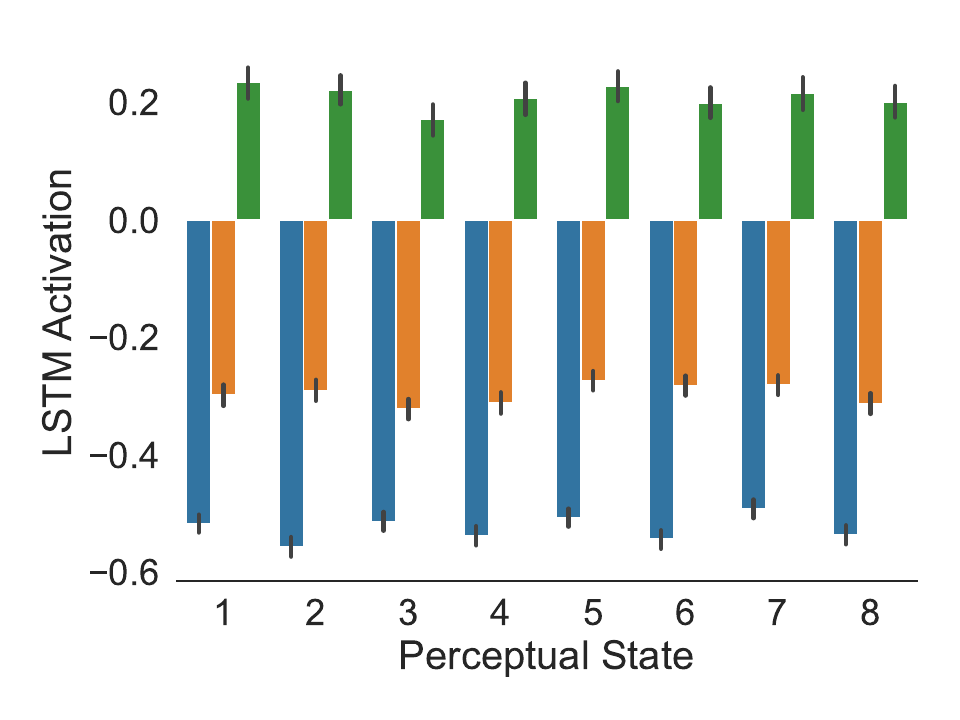}
        \end{subfigure}%
        \begin{subfigure}{0.25\textwidth}
        \caption{}
            \includegraphics[width=1\linewidth]{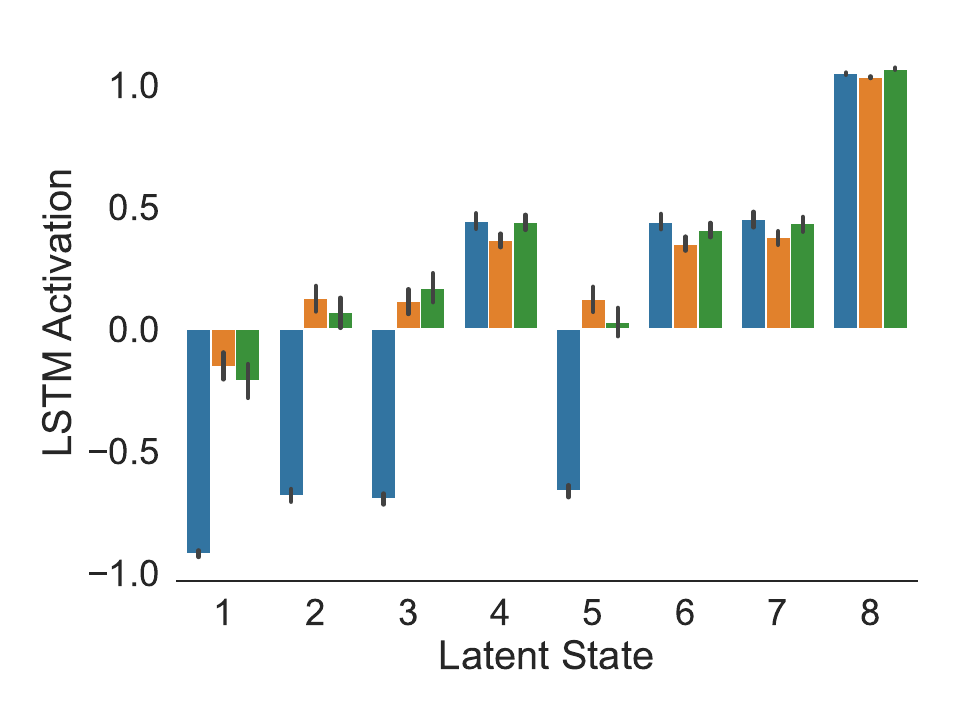}
        \end{subfigure}%
        \begin{subfigure}{0.25\textwidth}
        \caption{}
            \includegraphics[width=1\linewidth]{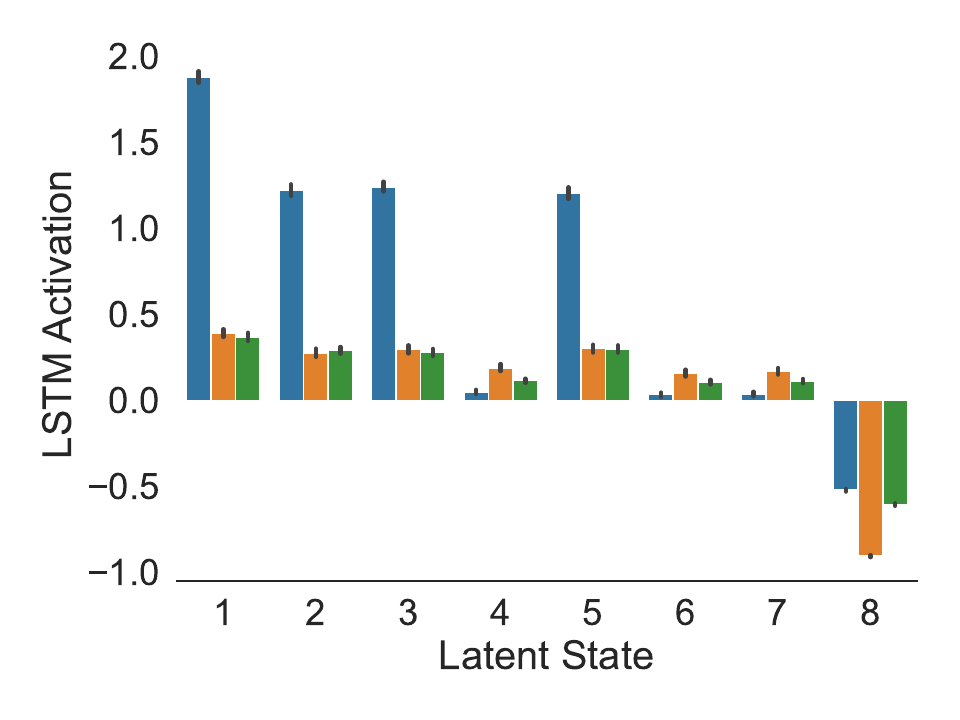}
        \end{subfigure}
        \begin{subfigure}{0.25\textwidth}
            \includegraphics[width=1\linewidth]{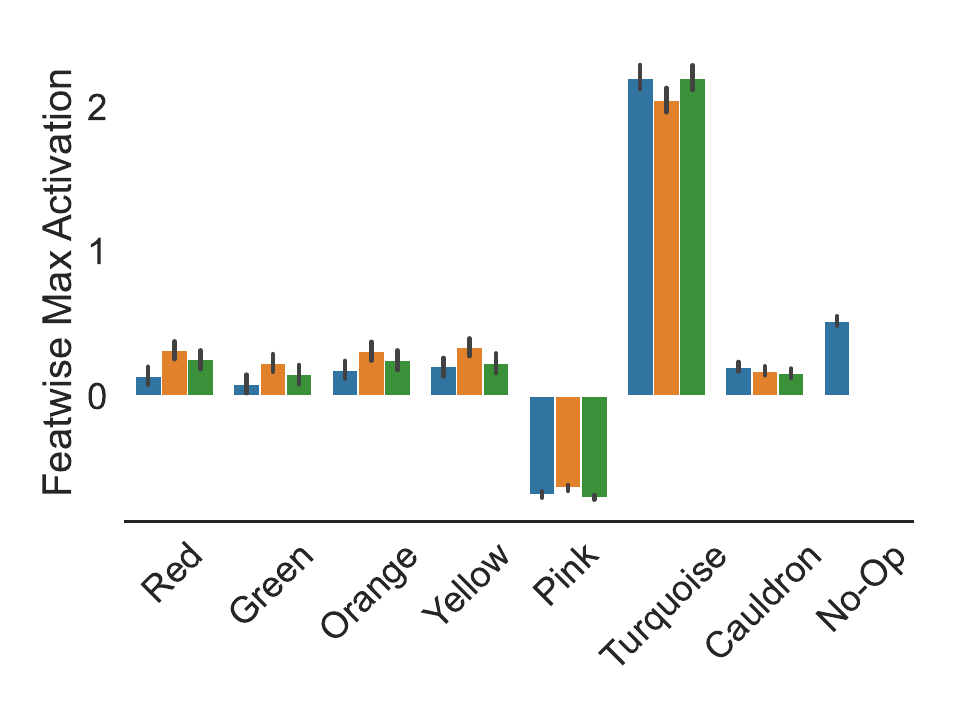}
        \end{subfigure}%
        \begin{subfigure}{0.25\textwidth}
            \includegraphics[width=1\linewidth]{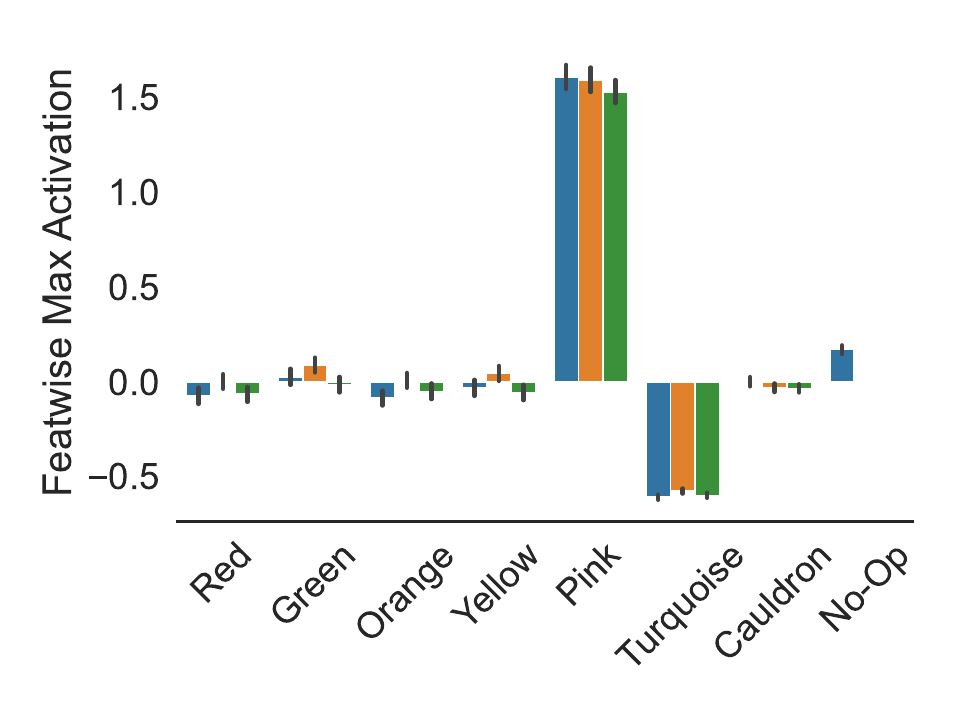}
        \end{subfigure}%
        \begin{subfigure}{0.25\textwidth}
            \includegraphics[width=1\linewidth]{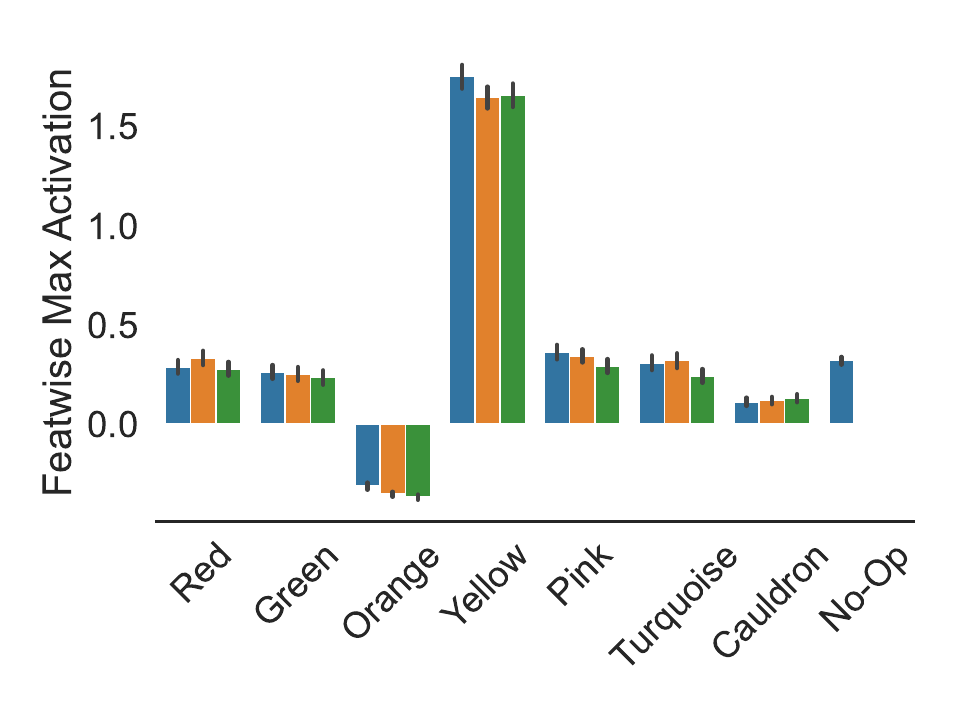}
        \end{subfigure}%
        \begin{subfigure}{0.25\textwidth}
            \includegraphics[width=1\linewidth]{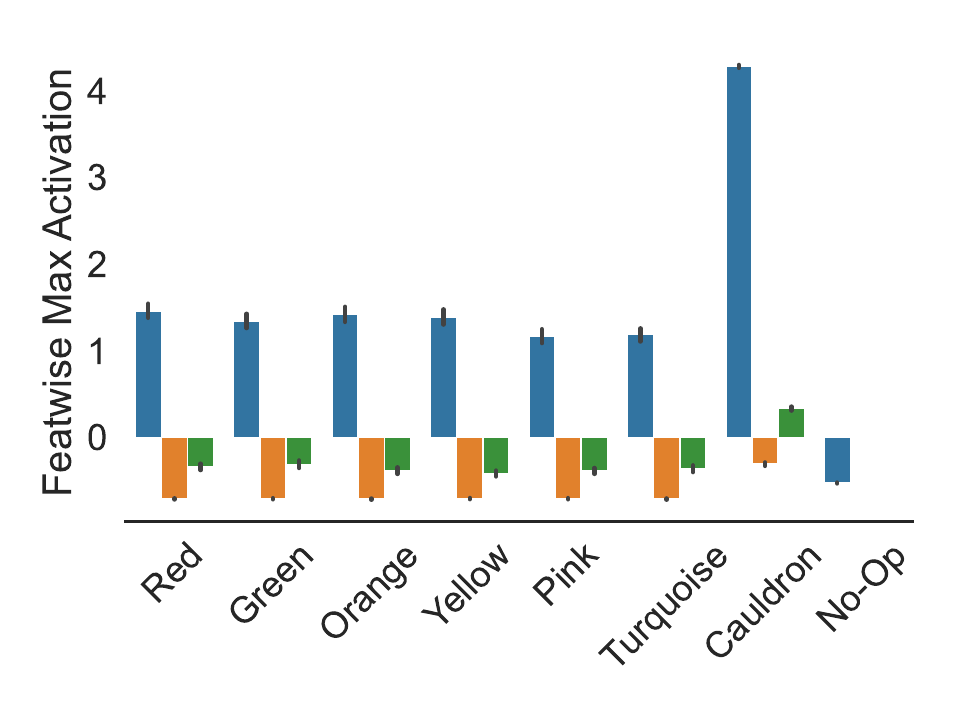}
        \end{subfigure}%
        \caption{\textbf{Top}: LSTM units. \textbf{(a)} The activation of unit $192$ as a function of the latent state. This unit is only responsive when stone $3$ is in latent state $8$ (the state with the highest reward). \textbf{(b)} The same unit as in (a) but as a function of the stone's perceptual state. It implies that the unit is responsive regardless of the stone's perceptual features. \textbf{(c)} The activations of unit $84$ has a magnitude proportional to the reward of its corresponding state regardless of which stone is used. Note that states $\{4, 6, 7\}$ have a reward of $+1$ while the rewards of states $\{2, 3, 5\}$ is $-1$. \textbf{(d)} The activations of unit $26$ are more positive in the states with negative reward. \textbf{Bottom}: Transformer units. \textbf{(a)} Unit $34$ has a high activation when the agent choose the turquoise potion while a negative activation when it chooses its opposite color. \textbf{(b)} Unit $56$ is the opposite of (a). \textbf{(c)} Unit $60$ is similarly responsive when the agent chooses potion with a specific color (here yellow) and has a negative response for its opposite color. \textbf{(d)} Unit 9 is selective when the agent chooses an action that uses stone 1 regardless of the potion color.}
        \label{fig:activations}
    \end{figure}
    
    Inspired by single-cell recordings in neuroscience, where single neurons are usually shown to be selective for a specific abstract concept \citep{wallis_single_neurons_2001}, we probed our model to see if it will give rise to similar selectivity by analysing the activations of single units in the LSTM and that of the transformer (specifically, the output of the feature-wise max). Figure \ref{fig:activations} shows the activations of a few units averaged across all steps in the $1000$ evaluation episodes. The latent and perceptual state of the stone that the agent chose is recorded along with the activation of each unit in each timestep. Note that there are $8$ latent states that correspond to the $8$ vertices of the cube, and $8$ possible perceptual states as previously mentioned in all of the held-out episodes. In the plots, we denote each state using a single number by binarizing its features. The top row shows the activations of some LSTM units while the bottom row shows activations recorded from the output the transformer. 
    
    Interestingly, we found distinct functional specialization between the transformer and the LSTM units. Specifically, $15.6\%$ of transformer units showed some understanding of the potion pairs abstract concept. Specifically, each of those units had a positive activation for one potion color and a negative activation for its opposite color (some samples shown in Figure \ref{fig:activations}), whereas no LSTM units did. The LSTM units, on the other hand, were mostly selective to stone-reward combinations. For instance, several units were only responsive when a specific set of stones had either a positive or negative reward and an opposite activation otherwise irrespective of the actual value or the stones' visual features. This also shows some notion of abstraction since the agent understands which stones it needs to deposit to the cauldron regardless of its exact representation.
    
    However, there were no single units that were responsive to a particular perceptual state or any single visual feature, nor were there units responsive to specific latent states except to the reward associated with that state. This led us to conjecture that the model reduces the cubic structure of the latent space to a two-dimensional form (i.e. a square), where each vertex correspond to one of the four possible rewards. This is inline with the observation that the agent is unable to handle bottlenecks in the underlying causal graph.

    
    
    
    

\section{Discussion and Future Work}

In this paper, we present an agent that is capable of meta-learning a set of abstract principles that underpins the complexity of the Alchemy benchmark. Further, we present a battery of analytical tools that can be used to test for specific pieces of abstract knowledge. Our design choices were motivated by the fact that strong deep-RL agents were unable to solve this task \citep{wang_alchemy_2021}, thus we gradually simplified the problem in order to identify exactly where the bottleneck lies. To that end, we reached some conclusions that can guide researchers in searching for better suited architectures. Specifically, our experiments show that more efficient outer-loop exploration and an architecture for handling positional output are required. We show as well that researchers with a limited compute budget can use Symbolic Alchemy in order to analyze their deep-RL agents in a principled manner. 

Finally, we found evidence for single-unit representations of abstract variables such as potion pairs, and functional dissociation between `cortical' and `hippocampal' units which were modeled as an LSTM and transformer respectively. This connects with what's been seen in recordings from rodent and primate cortex \citep{wallis_single_neurons_2001, samborska_complementary_2021, mansouri_review_2020}. This opens the way for future work to use meta-RL to carry out more detailed simulations of classic neuroscience experiments to better understand the mechanisms underlying the observed results.




\bibliographystyle{apalike}  
\bibliography{symbolic_alchemy}  

\appendix

\section{Appendix}

\subsection{Action Type for Ideal Observer and Random Heuristic}

    \begin{figure}[h]
        \centering
        \begin{subfigure}{0.5\textwidth}
             \includegraphics[width=1\linewidth]{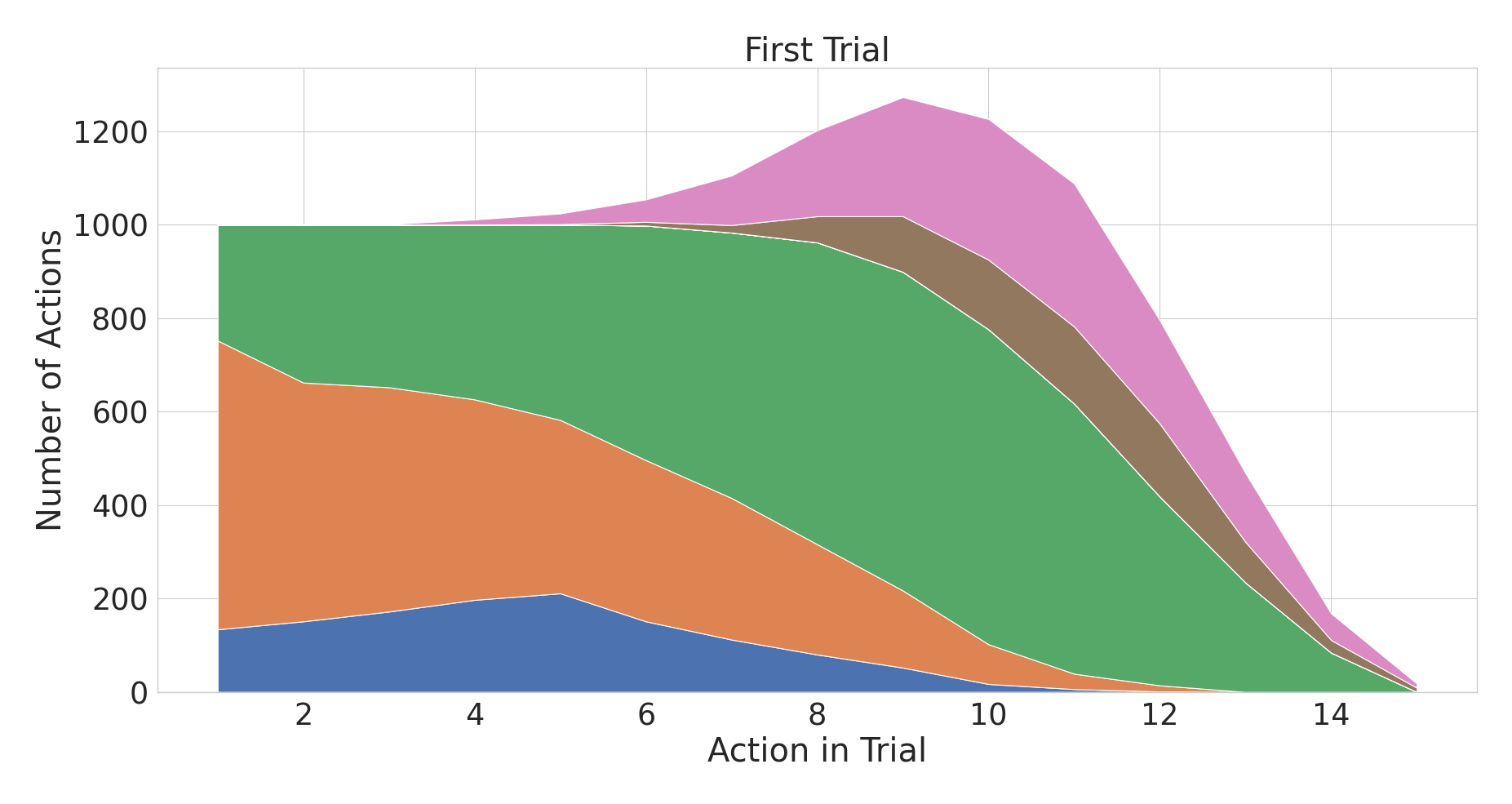}
        \end{subfigure}%
        \begin{subfigure}{0.5\textwidth}
             \includegraphics[width=1\linewidth]{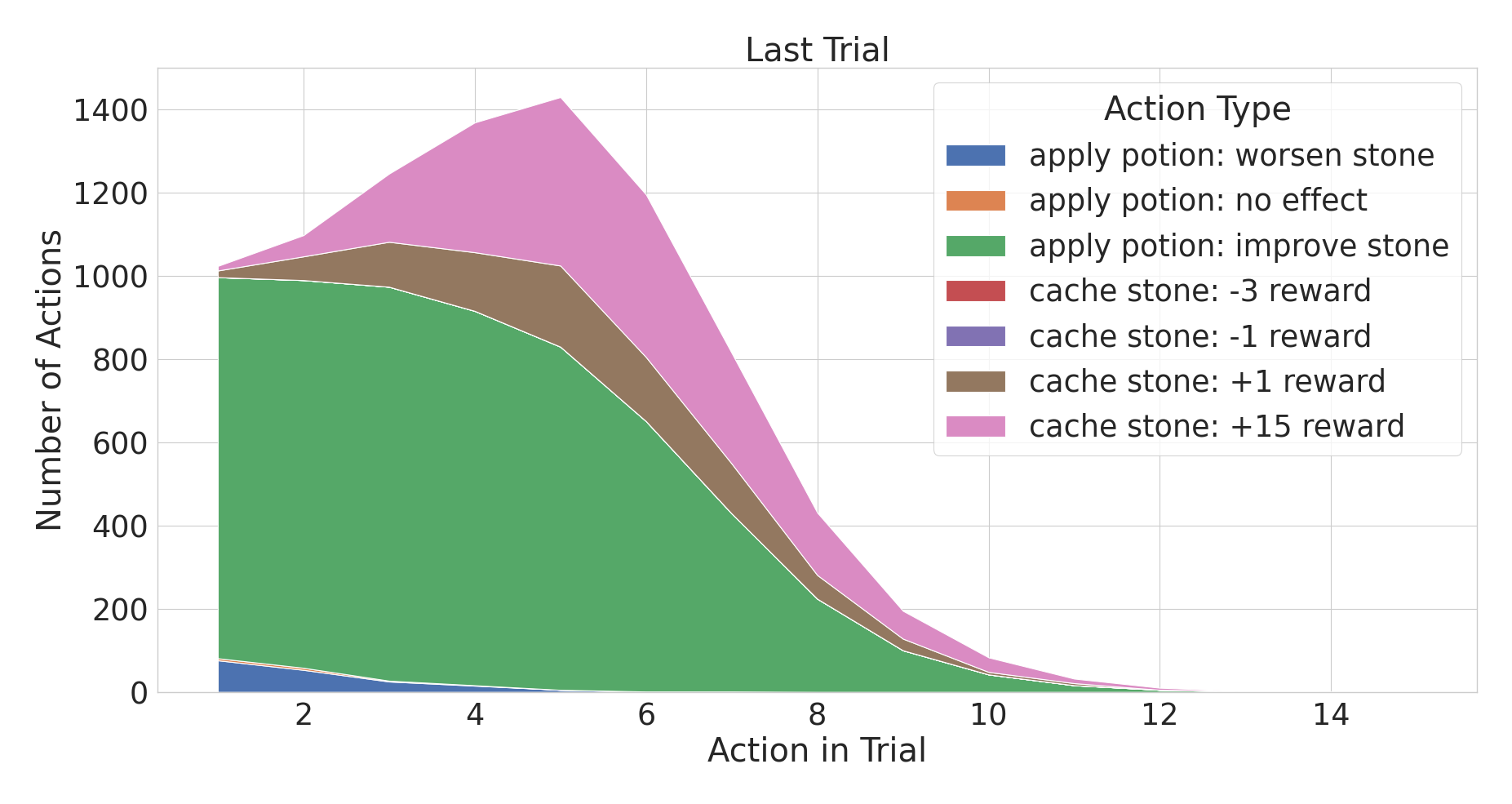}
        \end{subfigure}
        \caption{The type of action that the Ideal Observer take throughout the first and last trial. It can be seen that similar to the A2C EPN agent, it adapts its strategy across the episode.}
        \label{fig:io_action_type}
    \end{figure}
    
    \begin{figure}[h]
        \centering
        \begin{subfigure}{0.5\textwidth}
             \includegraphics[width=1\linewidth]{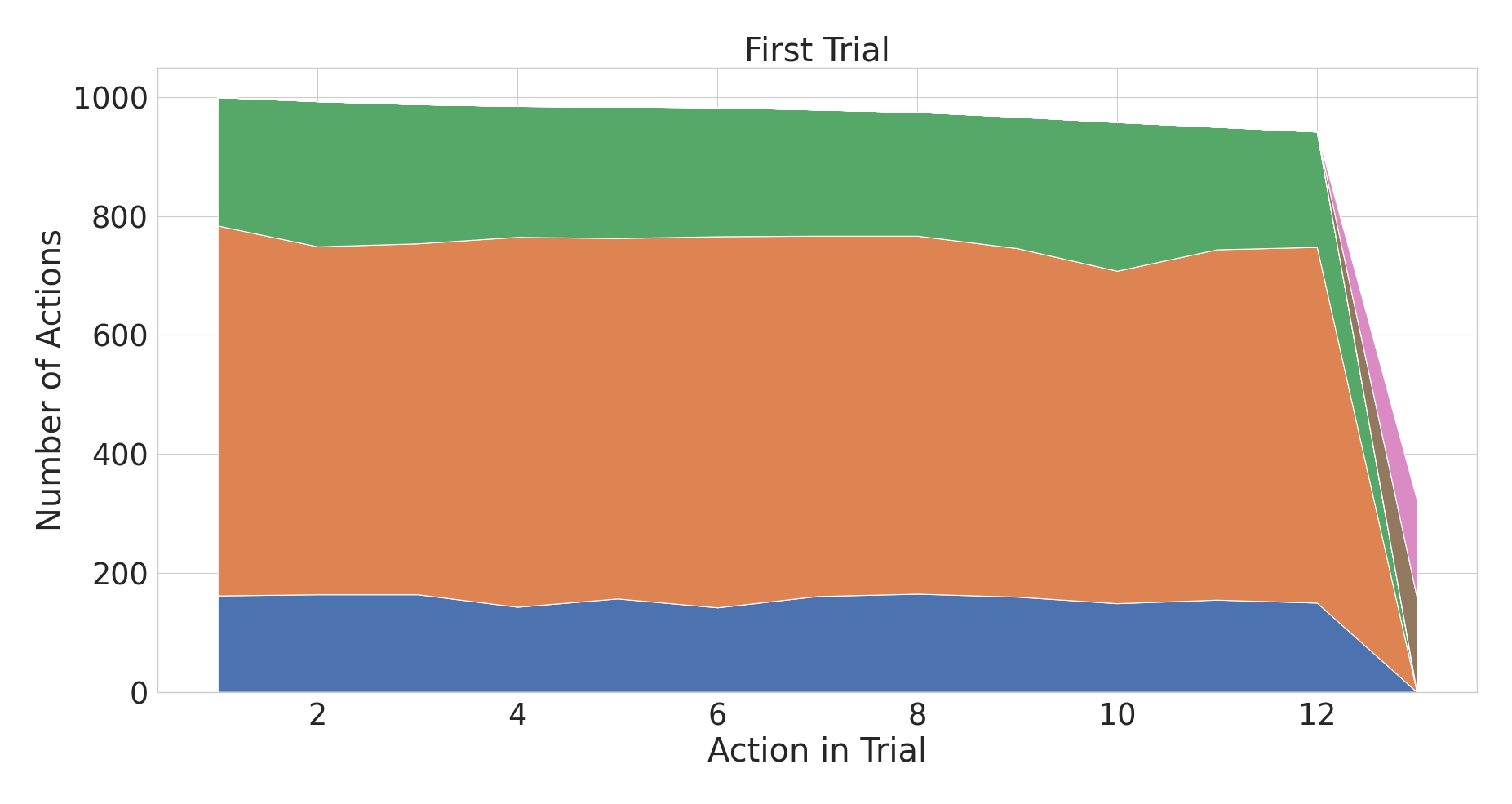}
        \end{subfigure}%
        \begin{subfigure}{0.5\textwidth}
             \includegraphics[width=1\linewidth]{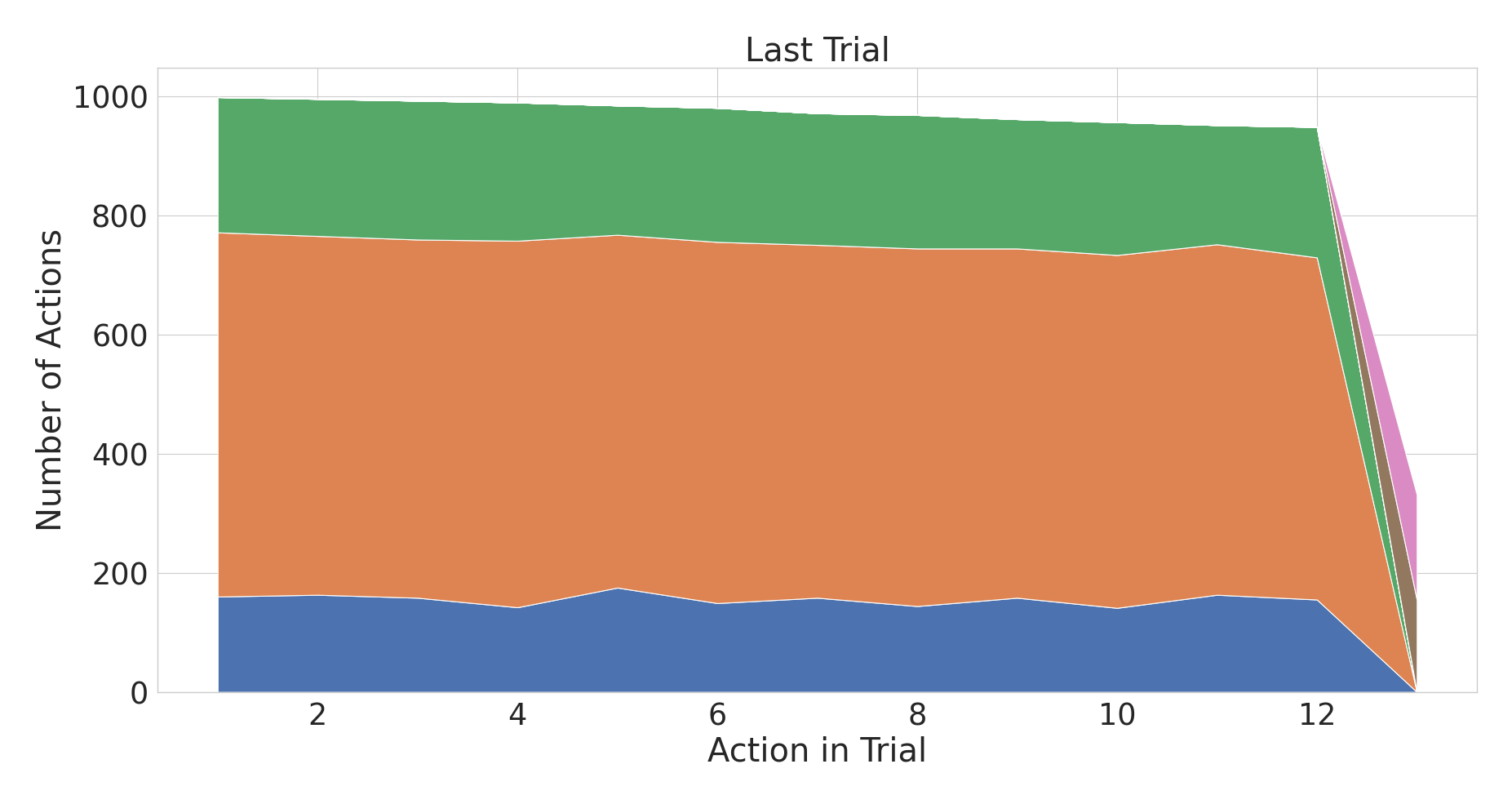}
        \end{subfigure}
        \caption{The type of action that the Random Heuristic take throughout the first and last trial. The agent's strategy is the same in both trials demonstrating an inability to adapt strategies.}
        \label{fig:randbot_action_type}
    \end{figure}

\subsection{Input, Output and Memory Representations}
\label{sec:io_explanation}

The canonical input and output representation are described in detail in \cite{wang_alchemy_2021}. The canonical memory representation on the other hand stores each entry as $[s_t, a_t, s_{t+1}]$ where $s_t$ is the current input state, $a_t$ is a one-hot encoding of the executed action at timestep $t$ and $s_{t+1}$ is the state at the following timestep that contains the transformed stone.

\end{document}